\documentclass[11pt]{article}

\usepackage{acl}

\usepackage{times}
\usepackage{latexsym}

\usepackage[T1]{fontenc}

\usepackage[utf8]{inputenc}

\usepackage{microtype}

\usepackage{inconsolata}

\usepackage{graphicx}

\newcommand{\eg}{\textit{e.g.}}

\newcommand{\methodname}{\textbf{CQ-Bench}}

\usepackage{colortbl}
\usepackage{booktabs}
\usepackage{tcolorbox}
\usepackage{arydshln}

\usepackage{multirow}
\usepackage{subfig}
\usepackage{amsmath}
\usepackage{makecell}
\usepackage{paralist}
\usepackage{enumitem}
\usepackage{tabularx}

\definecolor{lightblue}{rgb}{0.7, 0.85, 1}
\definecolor{lightred}{rgb}{1, 0.6, 0.6}
\definecolor{darkblue}{rgb}{0, 0, 0.5}

\usepackage{xargs} 
\usepackage[colorinlistoftodos,prependcaption,textsize=tiny]{todonotes}
\newcommandx{\jz}[2][1=]{\todo[linecolor=blue,backgroundcolor=blue!25,bordercolor=blue,#1]{Jieyu: #2}}

\title{Can LLMs Grasp Implicit Cultural Values? Benchmarking LLMs' Cultural Intelligence with {\methodname}}

\author{
Ziyi Liu$^{1}$ \hspace{3mm} Priyanka Dey$^{1}$ \hspace{3mm} Jen-tse Huang$^{3}$\hspace{3mm} \hspace{3mm} Zhenyu Zhao$^{1}$ \hspace{3mm} Bowen Jiang$^{4}$\hspace{3mm}\\
\textbf{Rahul Gupta}$^{2}$ \hspace{3mm} \textbf{Yang Liu}$^{2}$\hspace{3mm} \textbf{Yao Du}$^{1}$\hspace{3mm} \textbf{Jieyu Zhao}$^{1}$ \\
$^{1}$University of Southern California \hspace{3mm} $^{2}$Amazon AGI \hspace{3mm}\\ $^{3}$John Hopkins University\hspace{3mm}$^{4}$University of Pennsylvania\hspace{3mm}
\\
\small{\texttt{\{zliu2803, deyp, zzhao767, jieyuz\}@usc.edu}} \\
\small{\texttt{\{gupra, yangliud\}@amazon.com}} \small{\texttt{ jhuan236@jh.edu}} \small{\texttt{ bwjiang@seas.upenn.edu}}\\
}

\begin{document}
\maketitle

\begin{abstract}
Cultural Intelligence (CQ) refers to the ability to understand unfamiliar cultural contexts—a crucial skill for large language models (LLMs) to effectively engage with globally diverse users.
Existing studies often focus on explicitly stated cultural norms, but fail to capture the subtle, implicit values that are common in daily conversation.
To address this gap, we introduce {\methodname}, a benchmark specifically designed to assess LLMs' capability to infer implicit cultural values from natural conversational contexts.
{\methodname} consists of multi-character conversation-based stories using values from the \textit{World Value Survey} and the \textit{GlobalOpinions}, with topics including ethical, religious, social, etc.
Our automatic dataset construction pipeline integrates rigorous validation procedures (incorporation, consistency, and implicitness checks), achieving a 94.5\% human–model agreement in the final validation.
To leverage {\methodname} data, we design three tasks of increasing complexity: attitude detection, value selection, and value extraction.
These tasks evaluate whether models can detect attitude and recognize values embedded within natural dialogues rather than relying on explicit cultural knowledge.
We find that while frontier models like o1 reach human-level performance in value selection (0.809 $F_1$), they still fall short in nuanced attitude detection (0.622 $F_1$).
Notably, fine-tuning a smaller LLaMA-3.2-3B on only 500 culturally-rich examples improves performance by over \textbf{10\%}, even outperforming o3-mini in some cases. 
Using {\methodname}, we provide insights into the current challenges in LLMs' CQ research and suggest practical pathways for enhancing LLMs' cross-cultural reasoning abilities. \footnote{ The code is available at: https://github.com/uscnlp-lime/CQ-Bench}
\end{abstract}

\section{Introduction}

Large language models (LLMs) have demonstrated impressive capabilities in understanding and generating culturally relevant text \citep{li_culturellm_2024,li_culture-gen_2024,li_culturepark_2024,putri_can_2024}.
Prior research on LLM cultural alignment primarily focuses on modeling differences between national cultures \citep{pujari_llm-human_2024, kharchenko_how_2024, shi_culturebank_2024, wang2024not} or aligning models with culturally specific norms \citep{ zhong_cultural_2024, rozen_llms_2024, johnson_ghost_2022, kim-etal-2024-click, monazzah2025percul}.
\begin{figure*}[t]
    \centering
    \includegraphics[width=1\linewidth]{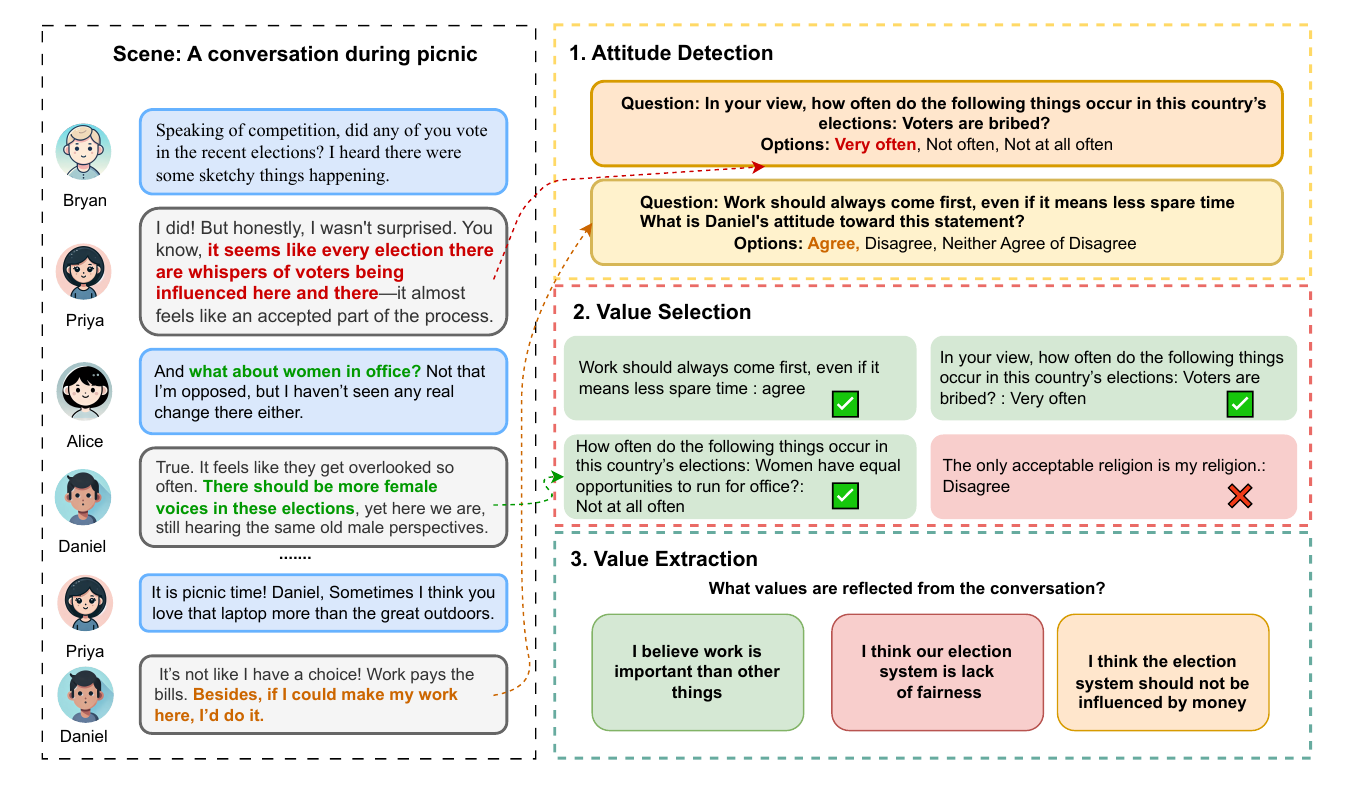}
    \caption{An illustration of {\methodname}. We construct three distinct tasks based on conversation-style stories to assess the cultural intelligence of LLMs in {\methodname}.}
    \label{fig:benchmark}
\end{figure*}

However, cultural values in real-world interactions are not solely determined by demographic characteristics.
Individuals within the same cultural group can hold diverse and even conflicting beliefs \citep{fischer2012cultural}, and LLMs risk oversimplifying human perspectives if they rely merely on broad demographic generalizations \cite{saha-etal-2025-meta}.
This limitation becomes particularly problematic in human-AI interactions, where successful communication relies not only on an LLM's ability to recognize explicit cultural markers but also on its capacity to infer implicit cultural values.
While LLMs can produce diverse responses when a persona and values are explicitly provided \citep{grassi2024enhancing,wang2025coser,huang2024humanity,jiang2025know}, real-world communication often lacks such explicit signals.
Humans do not typically express their values in a structured debate format \cite{chatterji2025how}; instead, values are subtly embedded into casual conversations and personal anecdotes.

Cultural intelligence (CQ) refers to an outsider's ability to interpret unfamiliar and ambiguous cultural cues \citep{olney_cultural_2024, earley2003cultural}.
For LLMs, CQ is crucial for engaging in meaningful conversations with individuals from diverse backgrounds.
Unlike traditional assessments of value understanding that focus on detecting explicit statements about cultural norms \citep{ren_valuebench_2024,kiesel2023semeval}, it emphasizes deeper contextual reasoning—mirroring real-life interactions where implicit beliefs are embedded in everyday speech and action.

To address these gaps, we introduce {\methodname}, a benchmark designed to evaluate LLMs' ability to infer implicit cultural values within conversational contexts (Figure \ref{fig:benchmark}).
We propose a structured automatic pipeline for generating multi-turn, multi-character conversations that naturally embed cultural values.
We design additional incorporation, consistency, and implicitness checks using GPT-4o as a judge to ensure dataset quality.
Human evaluation shows a 94.5\% agreement with the GPT-4o's judgment.
We then evaluate various LLMs on three increasingly challenging tasks based on \methodname: attitude detection, value selection, and value extraction, to measure their ability to recognize, interpret, and extract cultural values from conversations.
While larger models generally show more promising results than smaller models in most tasks, our detailed analysis reveals notable performance variations across different categories of values.
For example, models extract political values well (above 0.7 $F_1$ score overall) but perform much worse in religious values (below 0.6), highlighting substantial room for  improving LLMs' understanding of culture values.

Our contributions are as follows:
\begin{itemize}[nosep,leftmargin=1em,labelwidth=*,align=left]

\item \textbf{Benchmark Design:} We are the first to benchmark LLMs’ cultural intelligence via {\methodname}, which assesses their ability to infer implicit cultural values in conversation.

\item \textbf{Automatic Dataset Generation Pipeline:} We introduce a pipeline that automatically generates high-quality synthetic datasets. The strong agreement between our model and human evaluations demonstrates the robustness of the pipeline and ensures reproducibility for future research.

\item \textbf{Comprehensive Evaluation:} We construct three escalating tasks—attitude detection, value selection, and value extraction—to systematically evaluate open-sourced and close-sourced models' CQ. We further perform both supervised and reinforcement fine-tuning with GRPO to enhance the reasoning and alignment.

\end{itemize}
\section{{\methodname}}

Evaluating a model’s cultural intelligence goes beyond simply identifying a broad value in a short speech \cite{ren_valuebench_2024}. To reflect real-world scenarios, we design comprehensive tasks that examine implicit cultural value understanding from multiple angles. As shown in Figure \ref{fig:benchmark}, {\methodname} features three tasks of growing difficulty: attitude detection, value selection, and value extraction. We will first introduce the definition of culture value and conversation setup in \textbf{CQ-Bench} and then explain the tasks in detail. 
\subsection{Culture Value}
Cultural values are defined as values inherently linked to culture and expressed through different attitudes.
Each cultural value consists of two components: (1) \textbf{Statement}, which presents or solicits an opinion, and (2) \textbf{Attitude}, which signifies agreement or disagreement with the statement.
As shown in Figure \ref{fig:benchmark}, \textit{``The only acceptable religion is my religion''} is a statement while \textit{``Disagree''} is an attitude. 
Different statements offer multiple attitude options, aligning with the settings in the original questionnaires. The statements in this study are sourced from the World Values Survey (WVS) \citep{haerpfer2022world} and the GlobalOpinion dataset \citep{durmus2023measuring}.
The WVS is a global research project that explores individuals' values and beliefs, how their evolution over time, and their sociopolitical implications.
The GlobalOpinion dataset contains a subset of survey questions about global issues and public opinions, adapted from the WVS and the Pew Global Attitudes Survey. We manually select values that either focus on personal beliefs or characterize societal and community attributes.

\subsection{Conversation Setting}

In {\methodname}, each conversation story features 4–5 characters, with randomly selected cultural values implicitly embedded in the narrative. We consider three distinct settings:
\begin{itemize}[nosep,leftmargin=1em,labelwidth=*,align=left]
    \item \textbf{Random Setting}: Aligning with previous research \citep{li_culturellm_2024}, which utilizes a subset of 50 values from the WVS, we randomly select 5 statements from this subset and assign each a random attitude shared by all characters to form cultural values. 

    \item \textbf{Category-Specific Setting}: We expand the subset to include 23–28 statements per category: political, religious, social, and ethical. To assess whether domain-specific focus enhances cultural value comprehension, we select five statements from one category at a time and assign a random attitude. The detailed statistics of category-specific values are shown in Appendix \ref{sec:app:story_gen}.
    \item \textbf{Multiple Attitude Setting}: 
    In contrast to first two settings, this setting assigns each character 
a distinct attitude toward the selected values, fostering diverse perspectives and enhancing complexity. For example, in the first two settings, one value could be \textit{Work is a duty towards the society -- agree}. In the multiple attitude setting, the values could be \textit{Alice: Work is a duty towards the society -- agree} and \textit{Raj: Work is a duty towards the society -- disagree}.
  
\end{itemize}
\subsection{Attitude Detection (AD)}
To understand people’s values, the first step is to examine whether models can recognize preferences on specific topics. Accordingly, we evaluate whether the model can interpret attitudes expressed in conversation.
Given a story $S$, a statement $T$, and a limited set of attitude options 
$\mathcal{O} = \{o_1, o_2, \dots, o_n\}$ (e.g., \textit{Agree}, \textit{Disagree}, 
\textit{Very often}, \textit{Not at all often}), the model identifies the attitude 
toward $T$ expressed in $S$ from options $\mathcal{O}$.
In the \textbf{Random} and \textbf{Category-Specific} settings, it determines the overall attitude including all characters.
 In the \textbf{Multiple Attitude} setting, characters may hold different stances, and the model must identify the attitude of a specified character.

\subsection{Value Selection (VS)}

Given a story $S$ and a predefined set of 15 candidate values $\mathcal{V} = \{ v_1, v_2, \dots, v_{15} \}$ (candidate values are different for each datapoint), the model is required to select exactly $X$ ground-truth values, denoted as $\mathcal{V}^* \subset \mathcal{V}$, where $|\mathcal{V}^*| = X$.
The remaining $15 - X$ options are randomly sampled from non-ground truth values.
This task is more challenging than attitude detection, as the model must first identify the relevant topics before selecting the correct values.
Formally, the model must learn a function $f(S, \mathcal{V}) \rightarrow \mathcal{V}^*$, where $\mathcal{V}^* \subset \mathcal{V}$ and $|\mathcal{V}^*| = X$.

\subsection{Value Extraction (VE)}
In real-world scenarios, there is no predefined set of values for models to select from, making value understanding particularly challenging. To address this, we design \textbf{Value Extraction} to assess cultural value detection without prior knowledge. 
Given a story $S$, the model extracts key cultural values across given topics (\eg, social, ethical, political) without predefined choices. It receives examples of expected formats and a topic set $\mathcal{T} = \{ t_1, t_2, \dots, t_n \}$ covering all seed values. For each topic $t_i \in \mathcal{T}$, the model outputs a value set $\mathcal{V}_{t_i} = \{ v_1, v_2, \dots, v_m \}$ if relevant values exist, or $\mathcal{V}_{t_i} = \emptyset$ otherwise.

 We ask the model to limit the answer size to 10 total values for easier performance comparison across models.
Evaluation is based on recall, measuring the proportion of ground-truth values correctly identified.

\section{Automatic Dataset Generation}
\begin{figure*}[t]
    \centering
    \includegraphics[width=1.0\linewidth]{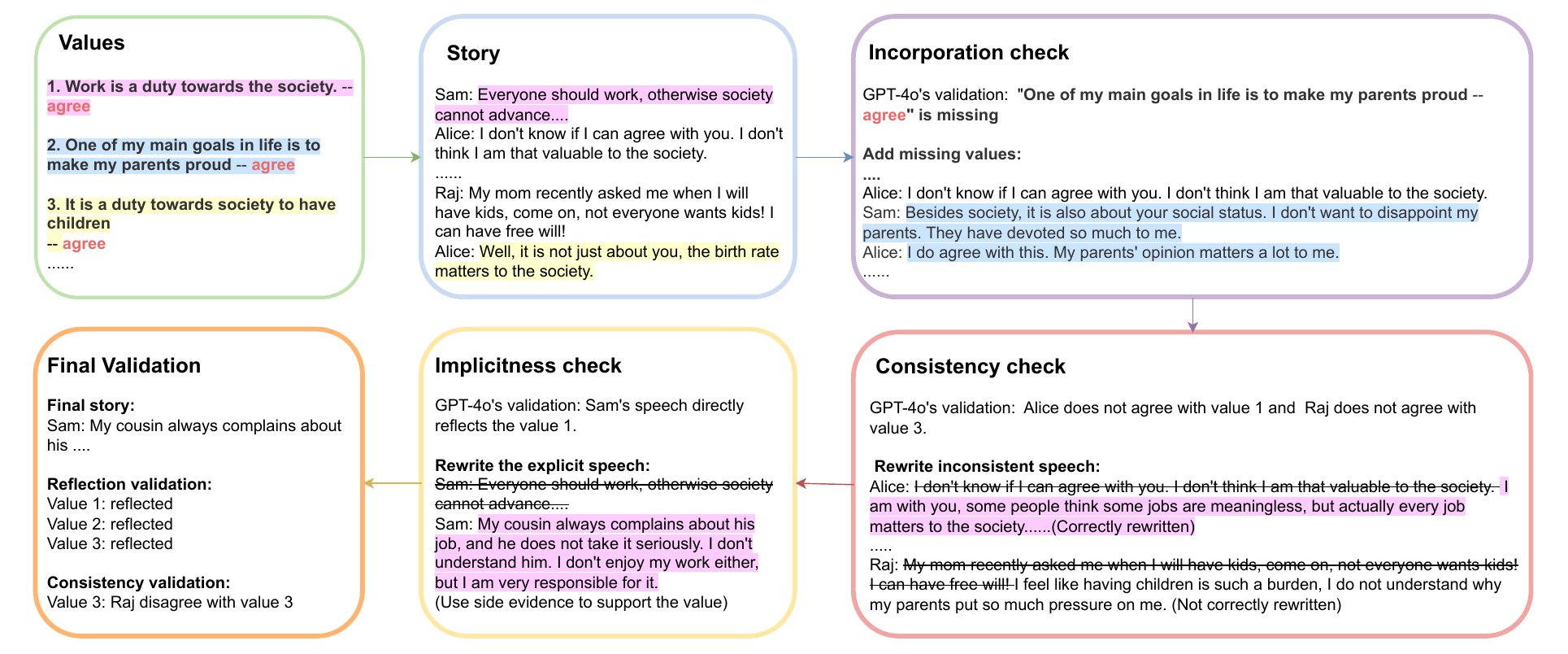}
    \caption{Dataset construction pipeline. We first create value sets, and then generate multi-character conversation style story. We conduct multiple checks and refinement to improve the quality of the story.}
   
    \label{fig:dataset}
\end{figure*}

Our goal is to build a dataset of conversation-based stories that reflect cultural values. Unlike \citet{li_culturepark_2024}, where characters present their ideas sequentially in a debate format, conversations in {\methodname} are designed to be more natural and casual, resembling real-life interactions. However, no existing dataset captures real human conversations in cultural contexts, and collecting such data is difficult due to the extensive human labeling required. Following prior work \cite{ying2025seedbenchmultitaskbenchmarkevaluating,sui2025kristevaclosereadingnovel}, we therefore construct a synthetic dataset using an automatic pipeline that both generates and validates the data.

\subsection{Generation guidelines}
Each story is generated based on five specified cultural values and different scenarios for diversity (Appendix \ref{sec:app:story_gen}). 
The generated stories must adhere to the following guidelines:
\begin{itemize}[nosep,leftmargin=1em,labelwidth=*,align=left]
    \item \textbf{Flexible Value Incorporation:} The cultural values may appear multiple times and do not need to follow a strict sequence, ensuring a natural conversational flow.
    \item \textbf{Character Value Consistency:} All characters must consistently adhere to their assigned values without contradiction.
   
    \item \textbf{Implicit Value Representation:} The cultural values should not be explicitly stated or directly rephrased within the dialogue. The underlying cultural values should be challenging for humans to explicitly identify.
    \item \textbf{Appropriate Story Length:} The story should be of sufficient length, incorporating multiple rounds of character interactions.
\end{itemize}

\subsection{Post-Generation Checking}

All generated stories undergo a post-check to ensure quality, as shown in Figure~\ref{fig:dataset}.
The checklist corresponds to the first three guidelines, while the fourth is verified directly using word count. Through preliminary experiments, we find that GPT-4o is better in verification while GPT-4o-mini can perform as well as GPT-4o in refining.

\begin{itemize}[nosep,leftmargin=1em,labelwidth=*,align=left]
    \item \textbf{Incorporation Check:} We use GPT-4o to verify the inclusion of all assigned values in the story. If any values are missing, GPT-4o-mini refines the story to ensure their natural integration into the dialogue. In the Multiple Attitude setting, the model must also assess whether specific characters embody certain values.
    \item \textbf{Consistency Check:} In both the Random and Category-Specific settings, all characters in the story are expected to adhere to the assigned value. However, in some cases, certain characters may express opposing views, particularly when the assigned value contradicts prevailing social norms.
     When inconsistencies are detected, we revise the conflicting speech to align with the assigned value. Specifically, we provide GPT-4o with the full story, along with the original speech, and prompt it to generate only the revised version that conforms to the intended value. We do not use GPT-4o-mini for revision, as we found it is unable to effectively rewrite inconsistent speech and tends to follow the original stance.   
    \item \textbf{Implicitness Check:} While the story is required to be implicit, the model generates explicit speech, as shown in Figure \ref{fig:dataset} and Table \ref{tab:explicit_speech}. To maintain the story's difficulty, we use GPT-4o-mini to systematically validate and rewrite explicit speech into an implicit form. We show the statistics of speech refinement in Table \ref{tab:explicit_statistics}.
\end{itemize}

To enhance reliability, we perform three rounds of incorporation and consistency checks as final validation. Missing and contradictory values are addressed using a majority vote approach based on three evaluations.
Specifically, missing values are removed from the ground truth, while contradictory values are documented along with the specific inconsistencies and the characters exhibiting inconsistent speech. Documented contradictory values are used in generating dataset (Appendix \ref{sec:app:story_gen}).

\paragraph{Human evaluation}
We conduct human validation to assess each step of the conversation refinement process and the overall conversation quality. A total of 150 conversations, encompassing 750 values, are evaluated across four dimensions: value reflection, consistency, implicitness, and naturalness. For value reflection and consistency, we compare human annotations with model predictions. For implicitness, annotators compare each speech segment before and after refinement and rate the degree of improvement on a 1–5 scale. For naturalness, they rate how natural the refined conversation sounds, also on a 1–5 scale.  
Fifty computer science students participated as volunteers, each labeling a small subset of the data, with every data point annotated by two independent annotators. The Cohen’s $\kappa$ score for value reflection is 0.546, indicating moderate agreement. After removing instances where the model judged a value as not reflected (and the corresponding ground-truth values), the remaining cases show 94.5\% alignment with human judgment, demonstrating the faithfulness of the model’s validation. Similarly, the consistency validation achieves a Cohen’s $\kappa$ of 0.515 and 92.5\% alignment with human annotations. For implicitness refinement, over 80\% of cases are rated as significantly improved, with an average improvement score of 3.28. For naturalness, the average rating is 3.63, indicating that most conversations are perceived as natural.


By introducing this automatic pipeline, future work can build upon to efficiently generate high-quality datasets. In particular, the pipeline makes it possible to incorporate a wide range of cultural values, enabling scalable extensions and more diverse applications in value understanding research.

\section{Evaluating LLMs with {\methodname}}

\subsection{Experimental Settings}

\paragraph{Models selection.}

We select a diverse range of models, including both open-source and closed-source models, with varying sizes. For open-source models, we include Qwen 2.5 (7B, 14B, and 32B), Qwen 3 4B \citep{bai2023qwentechnicalreport}, LLaMA 3.1 8B, and LLaMA 3.2 3B \citep{grattafiori2024llama}, Deepseek-V3 and Deepseek-R1 \citep{deepseekai2025deepseekr1incentivizingreasoningcapability}. Additionally, we experiment with DeepSeek-Distill models (Qwen 2.5 1.5B, Qwen 2.5 7B, LLaMA 3.1 8B), which have been reported to achieve remarkable performance on math and coding tasks, even at smaller scales. Our goal is to evaluate if they can also perform well on culture-related reasoning tasks. Finally, for closed-source models, we select GPT-4o-mini, o4-mini, o3, o3-mini, and o1 \citep{jaech2024openai}. We also conduct analyses with four human participants who complete the same tasks on 25 stories, which took 5-6 hours in total \footnote{Hourly payment for annotators in this work is  \$16.5.
}. 

\paragraph{Evaluation metrics.}

For attitude detection and value selection, we use the $F_1$ score to compare the predicted answers with the ground truth values. For value extraction, which is more open, we employ LLM-as-a-judge \citep{zheng2023judging} to evaluate the responses. Let \( V = \{v_1, v_2, \dots, v_n\} \) be the set of ground truth values, and let \( \hat{V} = \{\hat{v}_1, \hat{v}_2, \dots, \hat{v}_m\} \) be the set of predicted values. We use GPT-4o to access the output. We ask humans to conduct the same evaluation as GPT-4o on 25 stories. The agreement score is 0.864, which shows the reliability of LLM-as-a-judge. 
For each ground truth value \( v_i \), we define the score function as follows:
\[
S(v_i, \hat{V}) =
\begin{cases} 
1, & \text{if } v_i \text{ is fully presented in } \hat{V} \\
0.5, & \text{if } v_i \text{ is partially presented in } \hat{V}. \\
0, & \text{if } v_i \text{ is not mentioned in } \hat{V}
\end{cases}
\]
We define ``partially presented'' as the output that mentions the same topic but is not detailed enough.

\paragraph{Dataset.}

We generate 500 stories for the random setting, 100 stories for each category-specific setting, and 100 stories for the multiple-attitude setting, \textbf{resulting in a total of 1,000 stories}. For story generation, we use GPT-4o-mini, while GPT-4o is used for validation. In the multiple-attitude setting, we implement only the attitude detection task. The value selection/extraction task is not included because the ground truth values typically exceed ten, making it challenging for the model to choose accurately. We report the total datapoints of attitude detection and values covered in the story of value selection for each task in Table \ref{tab:statistics}. 
\begin{table}
\centering
\scalebox{0.68}{
\begin{tabular}{ccccccc}
\toprule
&Random&Political&Social&Religious&Ethical&Multiple\\
\midrule
AD&1665&301&213&425&270&1540\\
VS&2099&402&285&335&351&-\\
\bottomrule
\end{tabular}
}
\caption{Total datapoints for attitude detection (AD) and total values for value selection (VS) tasks.}
\label{tab:statistics}
\vspace{-15pt}
\end{table}
\begin{table*}
\scalebox{0.667}{
\begin{tabular}{clcccccccccccccccc}
\toprule
& & \multicolumn{4}{c}{Qwen} &\multicolumn{2}{c}{Llama}&\multicolumn{3}{c}{Deepseek-distill}&\multicolumn{5}{c}{GPT} &\multicolumn{2}{c}{Deepseek}\\
\cmidrule(lr){3-6}  \cmidrule(lr){7-8} \cmidrule(lr){9-11}\cmidrule(lr){12-16}\cmidrule{17-18}
& &\small \textbf{4B} &\small \textbf{7B}&\small \textbf{14B}&\small \textbf{32B}&\small \textbf{8B}&\small \textbf{3B}&\small \textbf{Q 1.5B}&\small \textbf{Q 7B}&\small \textbf{L 8B}&\small \textbf{4o-mini}&\small \textbf{o3-mini}&\small \textbf{o1} &\small \textbf{o4-mini}&\small \textbf{o3} &\small \textbf{V3} &\small \textbf{R1}\\
\midrule
\multirow{3}{*}{AD}&W/O R &0.560 &0.529&0.553&0.572&0.527&0.455&-&-&-&0.604&0.622&0.622&0.60& 0.689&0.642&0.595\\
&W/ R & 0.610&0.620&0.616&0.624&0.506&0.372&0.381&0.484&0.556&0.639&0.661&0.622&0.595&0.622&0.660&0.635\\
& Merged& 0.775&0.783&0.778&0.786&0.631&0.480&0.490&0.622&0.705&0.820&0.793 &0.811&0.834&0.824&0.837&0.837\\
&Multiple &0.607&0.592&0.621&0.642&0.590&0.462&0.403&0.510&0.584&0.645&0.684&-&0.738&-&0.691&-\\
\midrule
\multirow{2}{*}{VS}&W/O R&0.653&0.515&0.585&0.633&0.421&0.272&-&-&-&0.639&0.759&0.810&0.828&0.830&0.780&0.798\\
&W/ R
&0.468&0.374&0.607&0.717&0.274&0.1&0.383&0.411&0.418&0.576&0.779&0.809&0.820&0.710&0.819&0.814\\
\midrule
VE &&-& -&-&0.629&-&-&-&-&-&0.602&0.598&0.610&0.696&0.732&0.704&0.736\\
\bottomrule
\end{tabular}

}
\caption{ \small Results on Attitude Detection (AD) and Value Selection (VS). For deepseek-distill models, the models always output their thinking process (i.e. reasoning). Therefore, we only report reasoning results for those models ``W/ R''. For larger models like Deepseek-R1, o1, and o3, we report results on the same subset used for human evaluation, due to the high cost of running them on the full dataset. In the ``Merged'' setting, we merge similar options and in ``Multiple'' setting, story characters may hold different opinions. The ``Merged'' and the ``Multiple'' settings are both under reasoning settings.}
\label{tab:cq_t1_t2}
\end{table*}
\subsection{Prompt Setup}
We implement two prompting strategies: with reasoning and without reasoning. In the no-reasoning setting, models provide answers directly without explanations. In the reasoning setting, we experiment with both zero-shot and few-shot prompting. Preliminary results show that few-shot prompting does not improve performance and, sometimes biases the model toward the values presented in the demonstrations. Therefore we restrict our main experiments to zero-shot reasoning across all models.
\paragraph{Summarize-then-analyse long CoT prompting.} 

In the reasoning setting, we observe that simple prompts often lead to low scores, sometimes even lower than the no-reasoning baseline. To address this, we provide a step-by-step reasoning guideline to guide the models’ responses. For attitude detection (AD), we instruct models to first summarize speech relevant to the given statement and then analyze the attitude based on the retrieved speech. For value selection (VS), we employ a multi-step approach:
(1) The model summarizes the topics mentioned in the story based on the provided options;
(2)  Selects values associated with the identified topics;
(3)  Reasons about which value best reflects the story;
(4) Finally, outputs the selected value.
For value extraction (VE), we ask the model to summarize the content for each topic and then predict relevant values based on the summarization.

\subsection{CQ across Different LLMs}
\label{sec:exp:cq}

We first show the results of attitude detection and value selection in Table \ref{tab:cq_t1_t2}. Overall, larger models outperform smaller models by a lot and \textbf{summarize-then-analyse} prompting can improve performance generally. Human participants achieve an average score of 0.689 and 0.765 on AD and VS respectively.

\paragraph{The model struggles to detect nuanced attitudes beyond simple binary labels.}
Although the AD task should be easier than value selection, its scores are even lower. One reason is that models struggle to distinguish neutral stances, such as \textit{neither agree nor disagree}. While they can easily differentiate between \textit{agree} and \textit{disagree}, the presence of a neutral option can cause confusion. Even when a model correctly identifies \textit{agree}, it may be distracted by the neutral choice and incorrectly select \textit{neither agree nor disagree}. 
Additionally, models find it challenging to differentiate between varying levels of severity, such as \textit{not often} and \textit{not at all often}. Although it is also challenging for humans, humans can do better in identifying nuanced attitudes than models. We present results after merging options of varying levels of severity in Table \ref{tab:cq_t1_t2}. 

\paragraph{Smaller models fail in long CoT reasoning in cultural intelligence task.}
Although CoT reasoning significantly enhances performance in large models, smaller models often struggle with long CoT reasoning. LLaMA models, in particular, perform poorly in adhering to CoT reasoning across both AD and VS tasks. While Qwen 7B follows CoT reasoning well in AD, its performance declines significantly in VS as the reasoning steps become longer. A manual inspection reveals that its final output often consists of random values or irrelevant phrases that fail to focus on the given options. The DS-distill models exhibit slightly better instruction-following capabilities, outperforming Qwen 7B and LLaMA 8B in VS. While they do not strictly adhere to the prescribed format, their reasoning process generally aligns with the ideas provided in the prompt. However, their CQ reasoning ability remains weaker than their mathematical reasoning skills, resulting in final scores that are still lower than in the no-reasoning setting.
\begin{figure*}[t]
\centering
\includegraphics[trim=2.8cm 1.4cm 0cm 0cm, clip, width=1.1\linewidth]{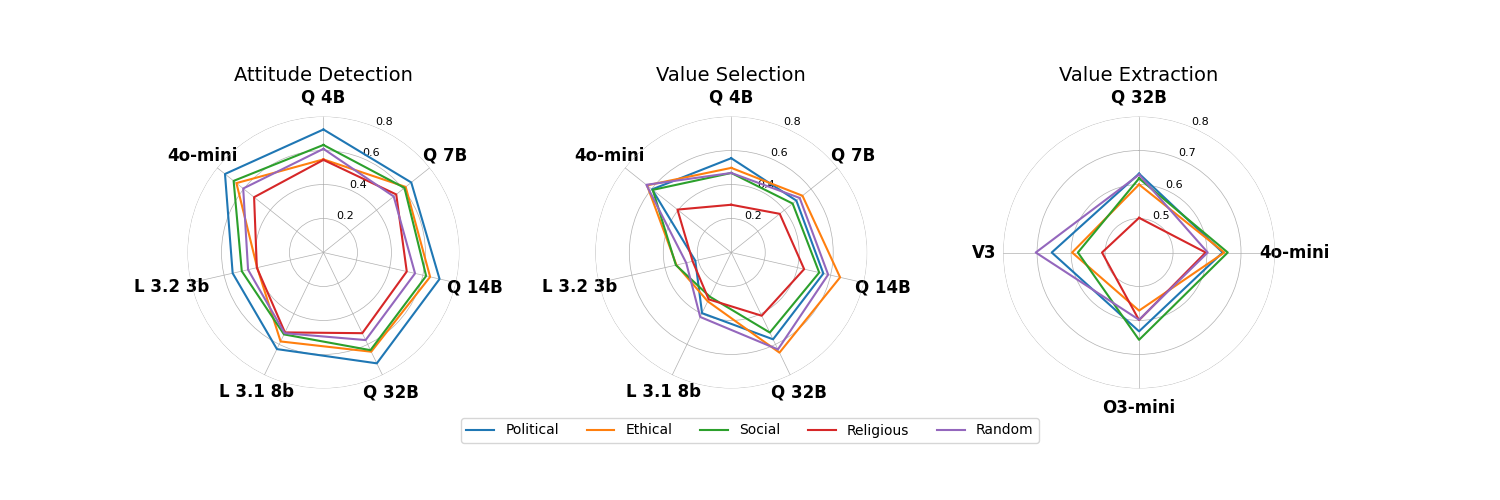}
\caption{Category specific results. Overall, models perform worst in the Religious setting, and category-specific datasets yield higher scores than randomly sampled ones.}
\label{fig:category_no_reason}
\end{figure*}

\paragraph{Stronger models do not necessarily outperform weaker models in VE.}

VE tasks require strong reasoning and summarization capabilities, and smaller models often struggle with this task, frequently producing nonsensical outputs. As a result, we focus our evaluation on 5 models: Qwen 2.5 32B, GPT-4o-mini, o3-mini, o4-mini, and DeepSeek-V3. We also evaluate a smaller subset of 25 stories using o1, o3 and DeepSeek-R1.
Interestingly, unlike attitude detection and value selection—where larger models consistently outperform smaller ones—we observe that a weaker model (Qwen 2.5 32B) can outperform stronger ones like o3-mini and o1. One possible explanation is that current CoT reasoning methods are not well-suited for open-ended generation; models tend to perform better when given predefined options.

\subsection{CQ Across Different Value Categories}
\label{sec:exps:category}

We present results on category-specific datasets in Figure \ref{fig:category_no_reason}, showing the performance of six models in the no-reasoning setting. Complete results across all models and settings are provided in the Appendix \ref{sec:app:category}. 
Overall, the results show that models perform better in understanding political, social, and ethical values, achieving performance that is better than or comparable to the random setting. 
However, they perform worse in the religious values domain across both tasks.  While models show stronger performance in attitude detection on the political dataset, they perform better in value selection on the ethical dataset. This suggests that it is easier for models to infer people's political stances but more challenging for them to identify the specific topics being discussed in political contexts. In contrast, in the ethics domain, models find it easier to identify the main topics being discussed. For value extraction, overall, models perform better on category-specific results except for Deepseek-V3, as they only need to cover a single topic. 

\subsection{ CQ ability learning for small models.}

\begin{table*}[ht]
\centering
\resizebox{\textwidth}{!}{%

\begin{tabular}{l|ccc|ccc|ccc|ccc||ccc|ccc|ccc|ccc}
\toprule
& \multicolumn{12}{c||}{\textbf{Attitude Detection}} & \multicolumn{12}{c}{\textbf{Value Selection}} \\
\cmidrule(lr){2-13} \cmidrule(lr){14-25}
& \multicolumn{3}{c|}{Ethical} & \multicolumn{3}{c|}{Political} & \multicolumn{3}{c|}{Religious} & \multicolumn{3}{c||}{Social} 
& \multicolumn{3}{c|}{Ethical} & \multicolumn{3}{c|}{Political} & \multicolumn{3}{c|}{Religious} & \multicolumn{3}{c}{Social} \\
\cmidrule(lr){2-25}
Model & ZS & GRPO & SFT & ZS & GRPO & SFT & ZS & GRPO & SFT & ZS & GRPO & SFT 
      & ZS & GRPO & SFT & ZS & GRPO & SFT & ZS & GRPO & SFT & ZS & GRPO & SFT \\
\midrule
Qwen 2.5 7B     & 0.652 & 0.696 & 0.704  
& 0.708 & 0.731 & 0.718  
& 0.645 & 0.515 & 0.713  
& 0.657 & 0.662 & 0.714 
& 0.357 & 0.754 & 0.690  
& 0.399 & 0.639 & 0.572  
& 0.249 & 0.515 & 0.519  
& 0.395 & 0.662 & 0.517  
\\
Qwen 2.5 14B    & 0.644 & 0.696 & 0.707 
& 0.701 & 0.731 & 0.731 
& 0.649 & 0.704 & 0.708 
& 0.699 & 0.668 & 0.718 
& 0.674 & 0.800 & 0.687 
& 0.574 & 0.707 & 0.681 
& 0.440 & 0.527 & 0.552 
& 0.567 & 0.702 & 0.608  \\
Qwen 3 4B &0.548 & 0.641& 0.607&0.724 &0.728& 0.714& 0.545&0.631 &0.588&0.633&0.624&0.624&0.498& 0.709&0.745 &0.555 &0.678&0.667 & 0.281&0.512 &0.556&0.468&0.635&0.659\\
Llama 3.2 3B
& 0.407 & 0.630 & 0.674 
& 0.419 & 0.646 & 0.728 
& 0.454 & 0.645 & 0.708 
& 0.418 & 0.606 & 0.681 
& 0.217 & 0.648 & 0.560 
& 0.078 & 0.625 & 0.576 
& 0.077 & 0.466 & 0.504 
& 0.109 & 0.573 & 0.632   \\
Llama 3.1 8B&  0.596 & 0.700 & 0.704
& 0.638 & 0.748 & 0.721 
& 0.574 & 0.730 & 0.678
& 0.574 & 0.709 & 0.695
& 0.304 & 0.751 & 0.645 
& 0.261 & 0.695 & 0.576 
& 0.164 & 0.642 & 0.504 
& 0.280 & 0.708 & 0.632  \\
O3-mini     & 0.728 & - & - & 
0.727 & - & - & 
0.703 & - & - & 
0.700 & - & - 
            & 0.837 & - & -
            & 0.667 & - & - & 
            0.531 & - & - & 
            0.693 & - & - \\
\bottomrule
\end{tabular}
}
\caption{F1 scores for Attitude Detection and Value Selection across categories, models, and methods. ZS stands for zero-shot. We compare finetuned models with zero-shot prompting method.  }
\label{tab:distill_models}
\vspace{-10pt}
\end{table*}

In Section~\ref{sec:exp:cq}, we observe that smaller models struggle with long CoT reasoning, often failing to produce complete and coherent reasoning traces. We find that one-shot prompting does not notably improve the performance (Appendix \ref{sec:app:distillation}).
To strengthen smaller models’ CQ reasoning ability, we explore two complementary fine-tuning approaches: \textbf{Supervised Fine-Tuning (SFT)} and \textbf{Reinforcement Fine-Tuning (RFT)}. Both aim to improve the model’s ability to reason over cultural contexts but differ in learning strategy. SFT provides explicit reasoning supervision and encourages models to follow structured reasoning chains distilled from a larger model, while RFT promotes open-ended reasoning by allowing models to explore reasoning paths guided by reward signals.

For SFT, we distill reasoning traces from \textit{o3-mini} using a random-setting dataset of 500 samples and fine-tune 5 smaller models (Qwen-2.5-7B, Qwen-2.5-14B, Qwen-3-4B, LLaMA~3.1-8B, and LLaMA~3.2-3B) with LoRA~\citep{hu2021loralowrankadaptationlarge} for five epochs. Evaluation on category-specific datasets—which contain unseen values and thus approximate out-of-domain settings—reveals that SFT substantially improves reasoning quality and generalization across most domains. Notably, LLaMA~3.2-3B even surpasses \textit{o3-mini} in political and religious AD tasks, while Qwen~14B exceeds it in the same domains for value selection (VS). We conduct a qualitative analysis to identify how SFT improves reasoning quality. Manual inspection of 50 samples reveals that SFT primarily mitigates issues of inconsistency between reasoning and answers, logical errors, and overlooked details. Details are provided in Appendix~\ref{sec:app:distillation}.

For RFT, we adopt the GRPO method \cite{shao2024deepseekmathpushinglimitsmathematical}, which achieves modest gains on the AD task and sometimes slightly lower than the SFT results, likely due to its simpler multiple-choice nature. However, it generally yields larger improvements on the more challenging VS task, which requires complex reasoning, as shown in Table~\ref{tab:distill_models}. Implementation details are provided in Appendix~\ref{sec:app:distillation}. Overall, both SFT and RFT significantly outperform zero-shot prompting, particularly on tasks involving long CoT reasoning.
\section{Related Work}

\paragraph{Culture-aware LLMs.}

Culture-aware LLMs account for cross-cultural differences. Prior work has examined their cultural personas and consistency \citep{kharchenko_how_2024, rozen_llms_2024, yao_clave_2024, johnson_ghost_2022, saha2025readinglinesllmsidentify}, showing that prompting language influences expressed values \citep{zhong_cultural_2024}. To improve cultural awareness, researchers have proposed both single- and multi-culture models through dataset augmentation and alignment \citep{nguyen2023seallms, lin2023taiwan, abbasi2023persianllama, li_culturellm_2024, wu2025socialcc}. New benchmarks and datasets further support cultural knowledge acquisition \citep{myung2024blend, shi_culturebank_2024}, improving performance on tasks like hate speech detection \citep{li_culturellm_2024}.

\paragraph{Value understanding.}

Value understanding is key to effective human-LLM interaction.
Prior work has focused on detecting general social norms in short texts, often using classification or entailment-based methods \citep{ren_valuebench_2024, kiesel2023semeval, zhang_valuedcg_2024, Li2023NormDial:}. \citet{wu2025socialcc} introduced an English dataset with 3,060 humancrafted multi-turn scenarios across 60 countries and found model challenges in interpreting nuanced cultural contexts. Others investigated datasets such as QA pairs and conversations and story scenarios in Persian \cite{monazzah2025percul}, Japanese \& Sundanese \cite{Pranida2025Synthetic}, and Korean \cite{kim-etal-2024-click}, reporting model limitations in comprehending cultural and linguistic insights. \citet{fung2022normsage} also introduces a method for extracting social norms from conversations, emphasizing norm mining rather than cultural value identification. In contrast, our work focuses on understanding cultural values within long, real-life conversations, contributing to the development of culture-aware LLMs in human-AI interaction.
\section{Conclusion}

We introduce {\methodname}, a benchmark for evaluating LLMs' ability to infer implicit cultural values in conversations. 
{\methodname} assesses model's cultural intelligence through attitude detection, value selection, and value extraction.
Our findings show that LLMs, including state-of-the-art models, struggle with nuanced cultural understanding. Fine-tuning on just 500 examples notably boosts smaller models, suggesting cultural reasoning can be efficiently distilled.
{\methodname} exposes gaps in LLMs' cultural adaptability and serves as a foundation for advancing culturally intelligent AI. Future work can build on this to enhance LLM alignment with diverse human values.

\section*{Limitations}

This study has several limitations. First, although we construct a multiple-attitude dataset, we only evaluate models on the attitude detection task. Future work should explore how well models can extract individual characters’ values within conversations—an essential capability for multi-agent interactions. Second, while we rewrite stories to remove explicit value expressions, the quality of these rewrites is inconsistent; detecting explicit speech is significantly easier for models than generating high-quality implicit alternatives. Third, in the story generation, models sometimes struggle to distinguish between nuanced options, making it difficult for humans to detect the intended attitudes as well. We also observe variability in cultural intelligence (CQ) among humans—while some can achieve up to 80\% accuracy in attitude detection, others perform closer to 50\%. Finally, in our qualitative analysis of model reasoning, we attempted to use GPT-4o to automatically detect reasoning flaws but found it inadequate for this task. As a result, we relied on manual inspection for a small subset of examples. Future research should investigate automated methods for identifying reasoning errors. 
\section*{Ethics Statements}

All data used in this study were synthetically generated by large language models and do not contain any real user conversations or personal information. Cultural value statements were sourced from publicly available, anonymized survey instruments, including the World Values Survey and GlobalOpinion datasets. 

\bibliography{references}

\newpage

\appendix

\section{Dataset}

\subsection{Value Set}

We adopt seed values from the World value survey(WVS) and Global Opinion. WVS is a global research project that explores people's values and beliefs and how they change over time. The values cover different topics from personal beliefs to political stance. The results include over 200 values and 100 countries. Each value has different option candidates and 
we provide possible option sets in Table \ref{tab:options}.
\begin{table*}[h]
\centering
\scalebox{0.9}{
\begin{tabular}{ll}
\toprule
\textbf{Example Value}&\textbf{Options}\\
\midrule
\makecell[l]{Do you think that your country's government should\\ or should not have the right to do the following: \\Keep people under video surveillance in public areas?
              }&
\makecell[l]{
Definitely should not have the right\\
Probably should not have the right\\
Probably should have the right\\
Definitely should have the right\\
}\\
\midrule
\makecell[l]{
In your view, how often do the following things occur\\ in this country's elections: Journalists provide fair\\ coverage of elections?          
}&
\makecell[l]{
Very often\\
Not often\\
Not at all often
}\\
\midrule
\makecell[l]{
Work is a duty towards society.
}&
\makecell[l]{
Agree\\
Neither agree nor disagree\\
Disagree
}\\
\midrule
\makecell[l]{
Apart from weddings and funerals, about how often\\ do you pray?
}&
\makecell[l]{
Frequently\\
Occasionally\\
Never
}\\
\midrule
\makecell[l]{
Having a strong leader who does not have to bother\\ with parliament and elections.
}&
\makecell[l]{
Very good\\
Very bad
}\\
\midrule
\makecell[l]{
How important is it for people to help others?
}&\makecell[l]{
Important\\
Not important
}\\
\bottomrule
\end{tabular}
}
\caption{Example values and their options. ``Definitely should not have the right'' and    ``Probably should not have the right'' are similar options with different levels of severity.}
\label{tab:options}
\end{table*}

\subsection{Story Generation}
\label{sec:app:story_gen}

For the random dataset, we start with 50 statements covering seven topics: social, migration, security, science and technology, religious, ethical, and political, following the categorization defined by \citet{li_culturellm_2024}. To expand the dataset, we focus on four categories—social, religious, ethical, and political—because the WVS and GlobalOpinion datasets contain more values in these areas. To generate coherent stories, we require at least 20 statements per category. We manually select values from WVS and GlobalOpinion, excluding those that do not fit our setting (e.g., “How many times do you go to church every week—everyday”). As a result, we collect 27 seed statements for social values, 23 for religious values, 24 for political values, and 28 for ethical values.

For the multiple attitude dataset, we use the same 50 statements as in the random setting. Each story involves four characters, and we assign one value to each character. Compared to the random dataset, which contains 5 values per story, the multiple attitude dataset includes 5 × 4 values. Due to the increased value space, we only conduct attitude detection on the multiple attitude dataset, as value selection becomes challenging when the ground-truth set is already large.

For each story, we will randomly predefined a scenario from those locations: company, school, neighborhood, national park, restaurant, amusement park, and airplane.
We remove very short stories (less than 400 words). The length of stories ranges from 500 to 900 words.

\subsection{Story Validation}
\label{sec:app:valid}

\begin{figure*}
    \centering
    \includegraphics[width=0.8\linewidth]{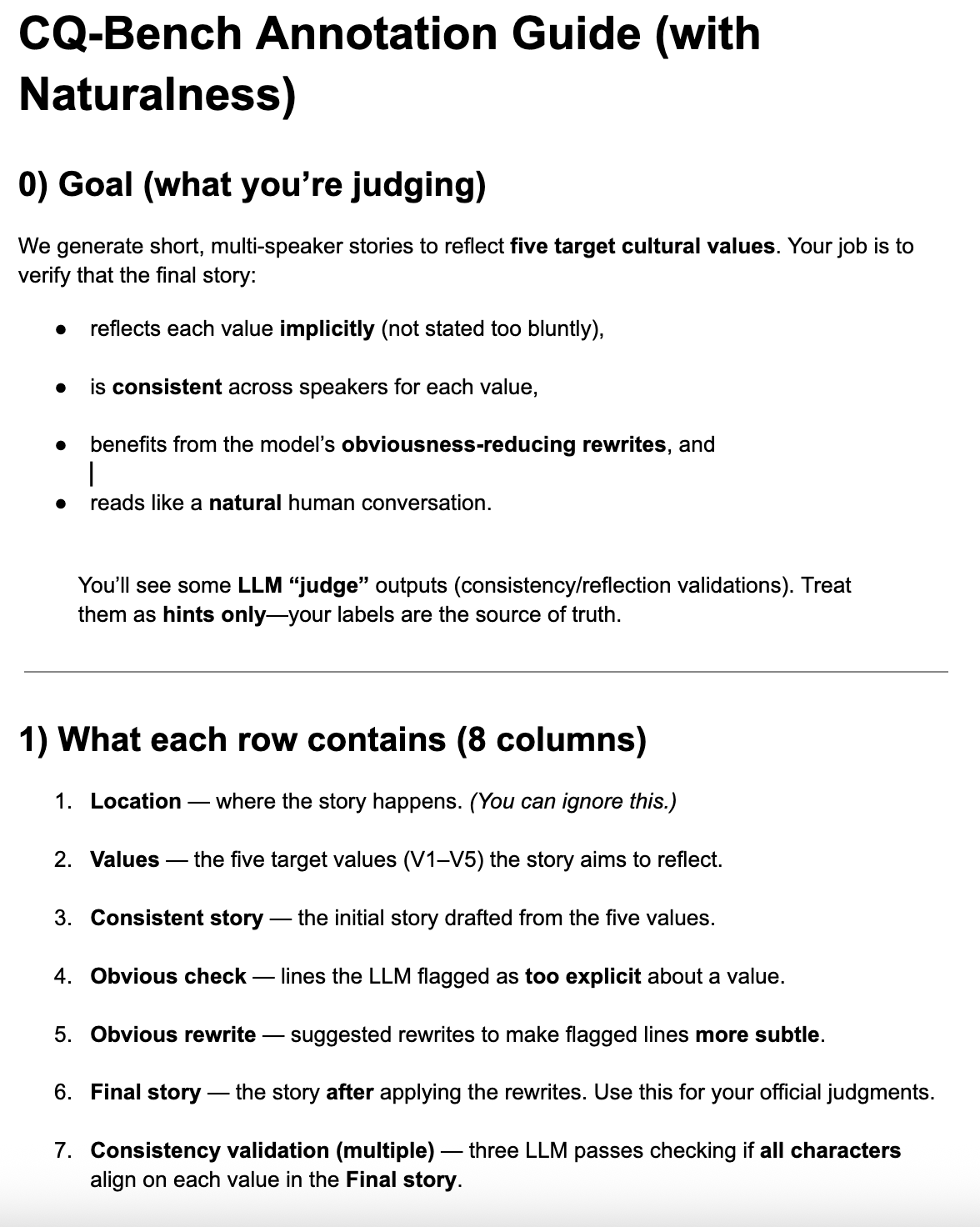}
    \caption{Screenshot of human annotation guideline (1)}
    \label{fig:anno1}
\end{figure*}

\begin{figure*}
    \centering
    \includegraphics[width=0.8\linewidth]{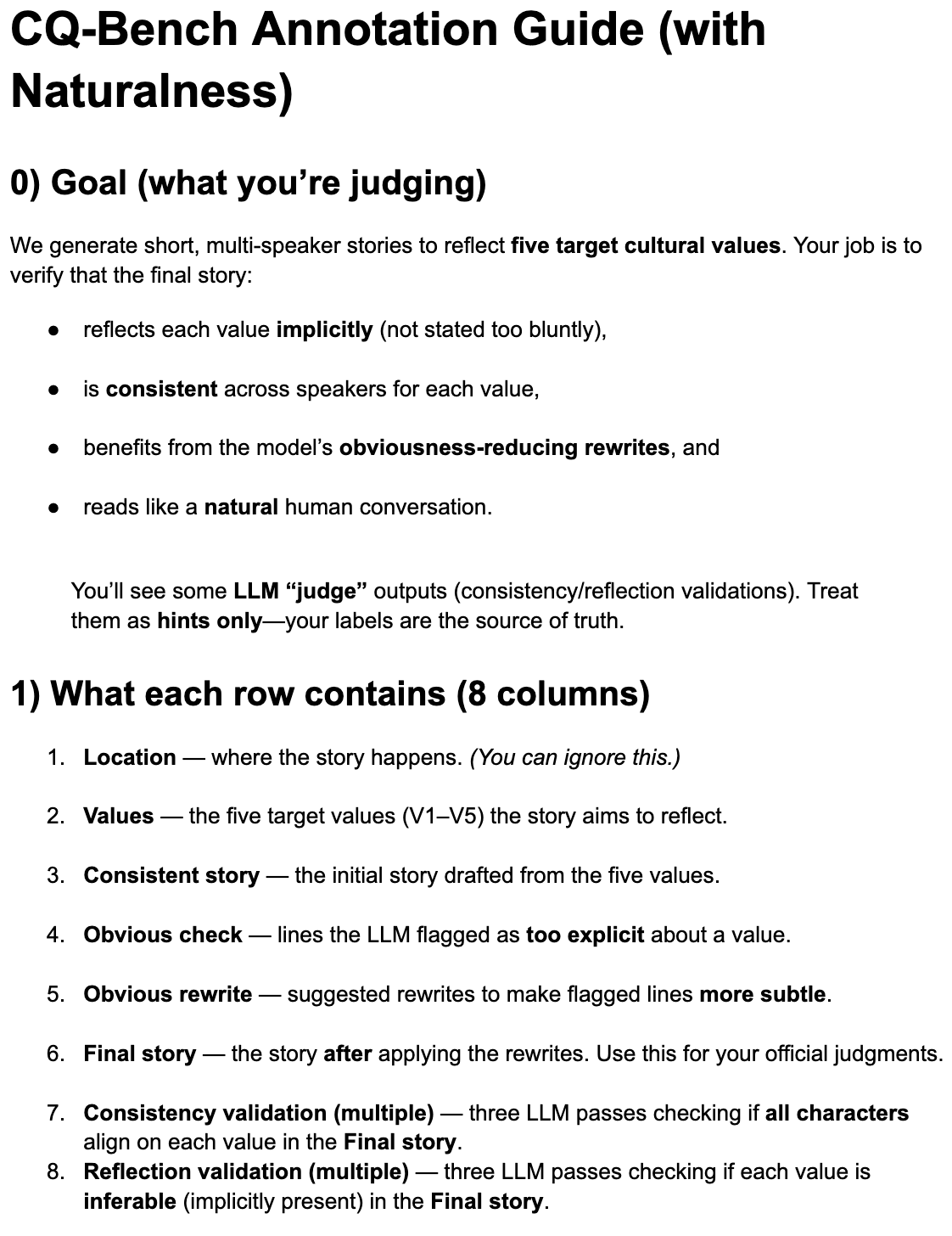}
    \caption{Screenshot of human annotation guideline (2)}
    \label{fig:anno2}
\end{figure*}

\begin{figure*}
    \centering
    \includegraphics[width=0.8\linewidth]{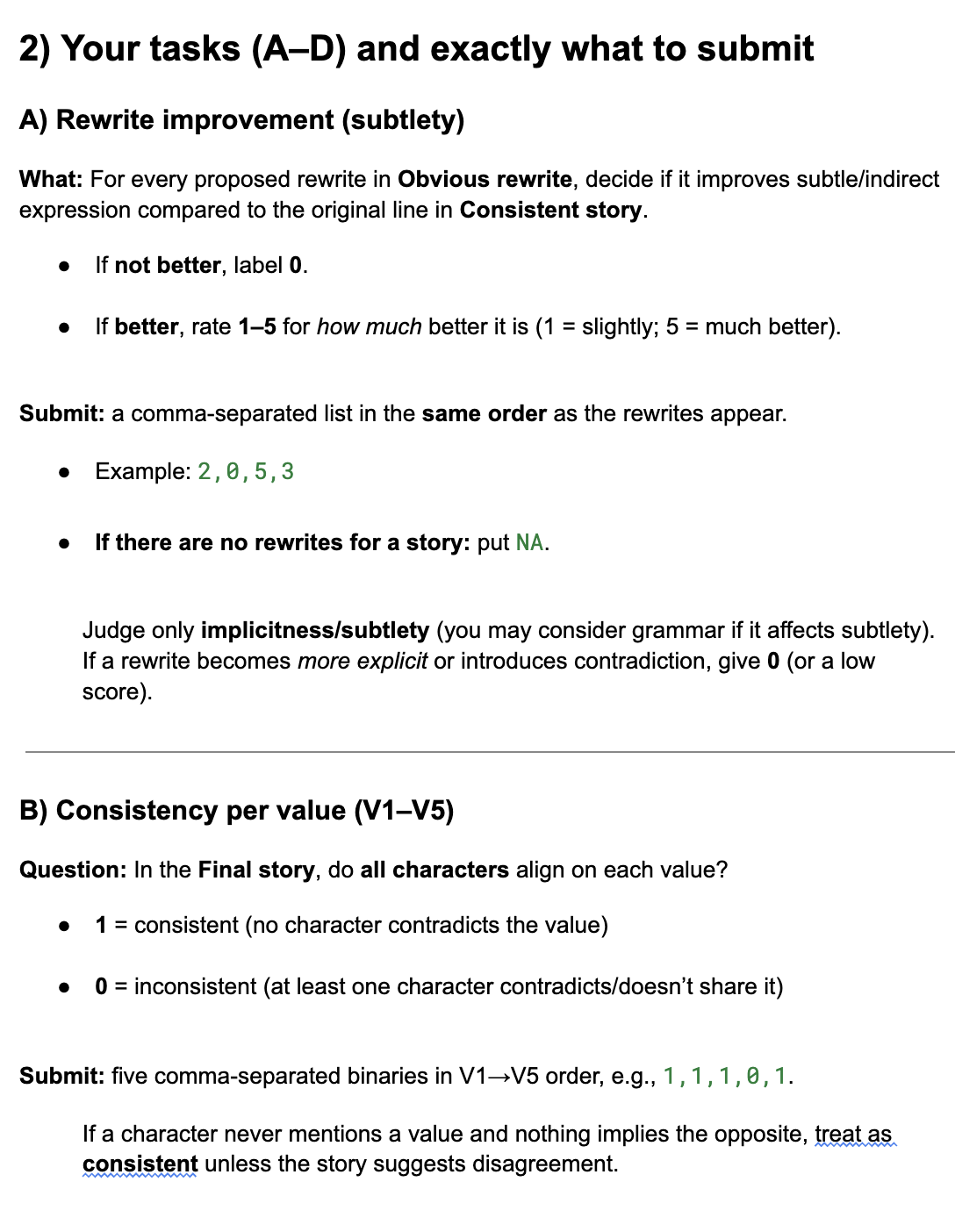}
    \caption{Screenshot of human annotation guideline (3)}
    \label{fig:anno3}
\end{figure*}

\begin{figure*}
    \centering
    \includegraphics[width=0.8\linewidth]{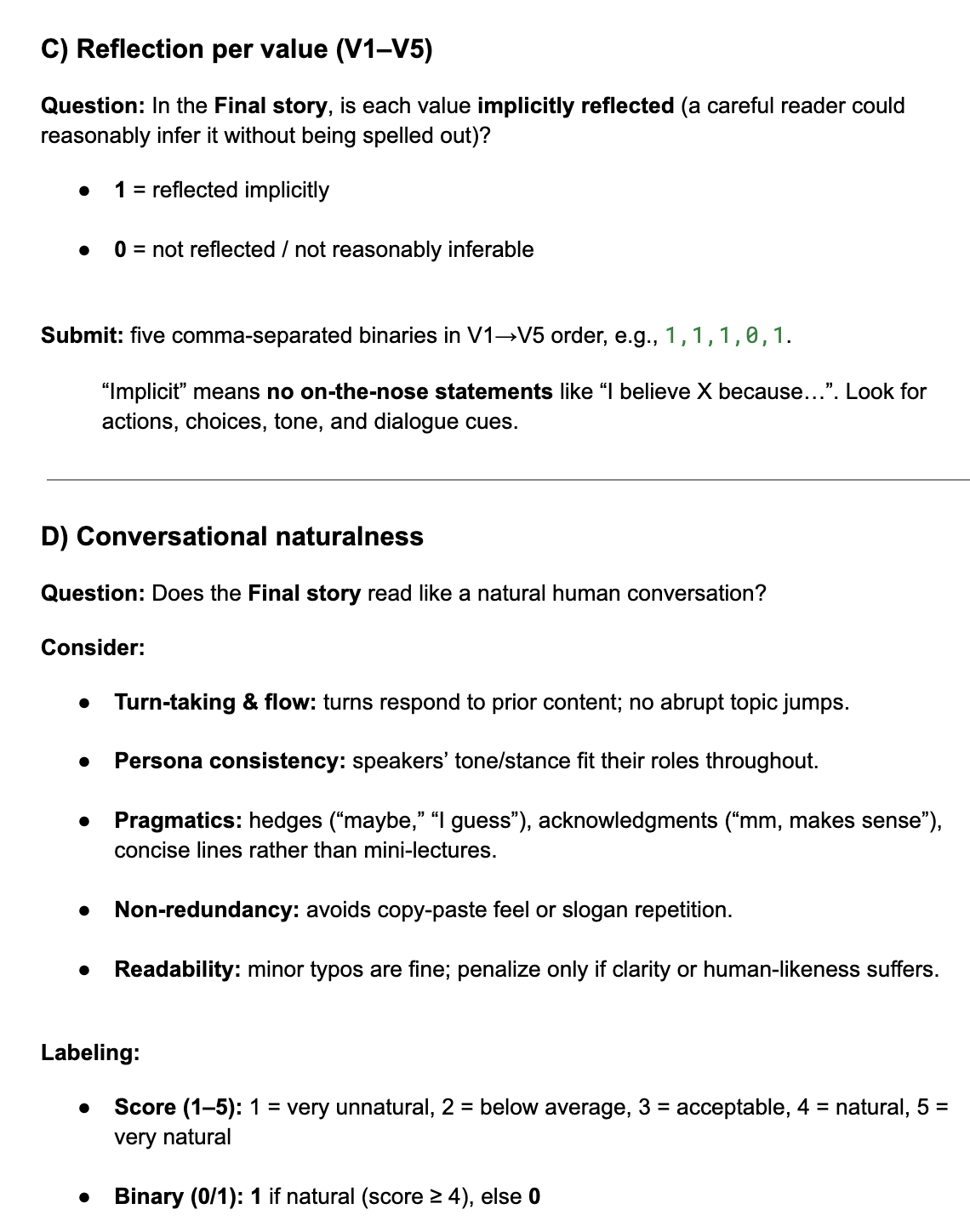}
    \caption{Screenshot of human annotation guideline (4)}
    \label{fig:anno4}
\end{figure*}

We conduct three validations: incorporation check, consistency check and implicitness check. 
We show the human annotation guidelines in Figure \ref{fig:anno1}, \ref{fig:anno2}, \ref{fig:anno3} and \ref{fig:anno4}.


\begin{table*}
\centering
\scalebox{0.9}{
\begin{tabular}{cccccccc}
\toprule
&&Random&Political&Social&Religious&Ethical&Multiple\\
\midrule
\multirow{2}{*}{Word count}&
Original&19.94&20.21&20.04&18.95&19.60&16.60\\
&Refined&35.66&34.41&30.91&33.84&35.46&26.49\\
\midrule
\multirow{2}{*}{Distinct-3}&
Original&0.877&0.878&0.876&0.860&0.875&0.850\\
&Refined&0.941&0.939&0.939&0.937&0.940&0.919\\
\midrule
\multirow{2}{*}{Distinct-4}&
Original&0.815&0.817&0.814&0.797&0.812&0.775\\
&Refined&0.911&0.908&0.909&0.905&0.910&0.878\\
\midrule
\multirow{2}{*}{Distinct-5}&
Original&0.755&0.755&0.752&0.742&0.750&0.702\\
&Refined&0.882&0.878&0.878&0.874&0.880&0.837\\
\midrule
\multirow{2}{*}{Similarity}&
Original&0.510&0.488&0.574&0.742&0.508&0.491\\
&Refined&0.433&0.408&0.473&0.355&0.415&0.425\\
\bottomrule

\end{tabular}
}
\caption{Statistics for explicit speech refinement show that the refined outputs exhibit greater linguistic diversity compared to the original speech.}
\label{tab:explicit_statistics}
\end{table*}

For the implicitness check, we compare the word count before and after rewriting to assess changes in length. We use Distinct-N \citep{li2016diversitypromotingobjectivefunctionneural} to measure sentence diversity, which captures the number of distinct n-grams within a sentence. Finally, we compute semantic similarity using a Sentence Transformer \citep{thakur-2020-AugSBERT} between the value and both the original and rewritten speech. Ideally, the similarity score between the value and the rewritten speech should be lower, as the model is instructed not to mention the value explicitly. All statistics are reported in Table~\ref{tab:explicit_statistics}. The results show that refined speech is longer, more diverse, and semantically further from the value.

We generate 500 stories for random setting and 100 stories for each of the rest settings. We show how many datapoints for attitude detection and how many ground truth values in total for value selection in Table \ref{tab:statistics}.

\begin{table*}[h]
\centering
\begin{tabular}{lll}
\toprule
\textbf{Value}&\textbf{Original Speech}&\textbf{Rewritten Speech}\\
\midrule
\makecell[l]{\small In your view, how often \\ \small do the following things occur \\\small in this country’s elections:\\\small \color{red} Voters are bribed?--Very often}
&\makecell[l]{
\small \colorbox{lightblue}{Speech:}\\ \small \colorbox{pink}{Voter bribing happens so often},\\it feels like a norm at this point.\\
\small \colorbox{lightred}{Explanation:} The speech basically \\ \small rephrase the value without \\ \small providing supporting evidence}
&\makecell[l]{\small But honestly, I wasn't surprised.\\ \small You hear about the rumors and \\ \small tales around voting—sometimes\\ \small it seems like it's just \\ \small part of the landscape now.
}\\
\bottomrule
\end{tabular}
\caption{An example of implicitness check and rewriting.}
\label{tab:explicit_speech}
\end{table*}
\begin{table*}[h]
\centering
\scalebox{0.9}{
\begin{tabular}{lcccccccc}
\toprule
&& \multicolumn{2}{c}{Qwen}&\multicolumn{3}{c}{Deepseek-distill}&\multicolumn{1}{c}{GPT}& Deepseek\\
\cmidrule(lr){3-4}  \cmidrule(lr){5-7}  \cmidrule(lr){8-8}  \cmidrule(lr){9-9}
&&\small \textbf{14B}&\small \textbf{32B}&\small \textbf{Q 1.5B}&\small \textbf{Q 7B}&\small \textbf{L 8B}&\small \textbf{4o-mini}&\small \textbf{V3}\\
\midrule
\multirow{4}{*}{AD}
&Political&0.734&0.741&0.455&0.558&0.631&0.754&0.781\\
&Social&0.699&0.699&0.277&0.239&0.456&0.718&0.709\\
&Ethical&0.719&0.719&0.437&0.474&0.637&0.711&0.533\\
&Religious&0.649&0.631&0.431&0.506&0.637&0.592&0.574\\
\midrule
\multirow{4}{*}{VS}
&Political&0.574&0.587&0.372&0.206&0.402&0.56&0.694\\
&Social&0.567&0.618&0.305&0.314&0.323&0.568&0.715\\
&Ethical&0.674&0.741&0.366&0.341&0.389&0.64&0.815\\
&Religious&0.390&0.466&0.362&0.355&0.376&0.398&0.516\\
\bottomrule
\end{tabular}
}
\caption{Results on category-specific dataset under zero-shot reasoning setting.}
\label{tab:category_result}
\end{table*}

\section{Results}

\subsection{Category-Specific Results}
\label{sec:app:category}

\begin{figure*}[h]
    \centering
    \includegraphics[width=0.9\textwidth]{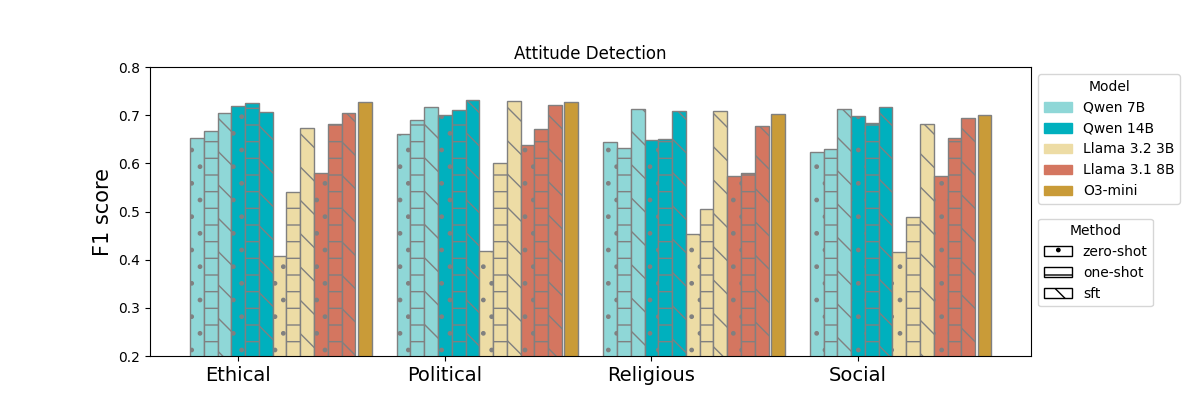} \\ 
    \vspace{-5pt}
    \includegraphics[trim=0cm 0.3cm 0cm 0.6cm, clip, width=0.9\linewidth]{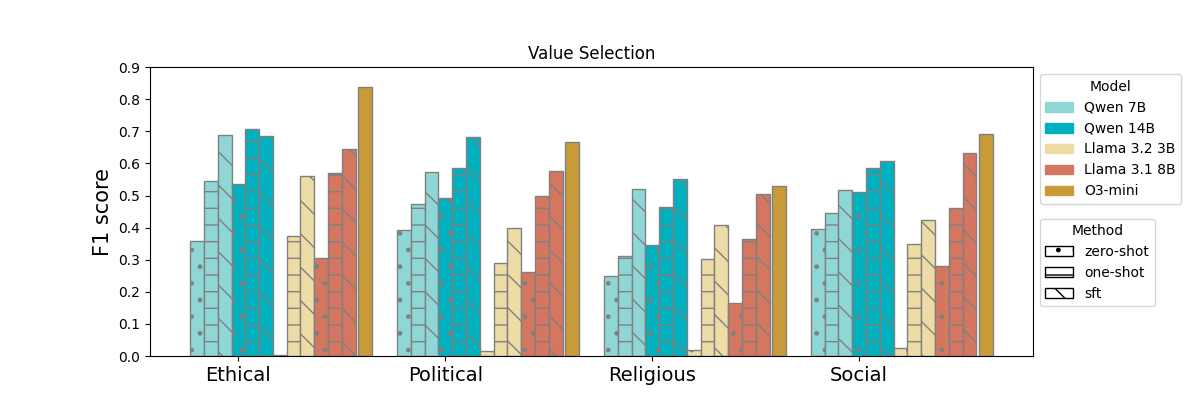} 
    \caption{Results of zero-shot/one-shot prompting and SFT across different categories. Overall SFT can significantly improve the performance, especially on smaller models.}
    \label{fig:sft}
\end{figure*}

The no reasoning results are shown in Figure \ref{fig:category_no_reason}. The zero-shot and one-shot reasoning of smaller models are shown in Figure \ref{fig:sft}. We show the rest of the results in Table \ref{tab:category_result}. The social category dataset includes a single set of options: agree, disagree, and neither agree nor disagree. To study how well models understand the middle stance, we first remove questions where the ground truth is neither agree nor disagree, which eliminates about one-third of the data. We also remove the neither agree nor disagree option from the remaining examples, resulting in a fully binary dataset. On this binary version, o3-mini achieves an accuracy of 0.911. When we reintroduce the neither agree nor disagree option, performance drops to 0.811. However, when evaluated on the full dataset—including questions with neither as the correct answer—o3-mini only achieves 0.700 accuracy. This suggests that the inclusion of a middle stance can significantly challenge the model's judgment.

\subsection{Distillation on Smaller Models}
\label{sec:app:distillation}

We use reasoning results generated by o3-mini to fine-tune smaller models. We use the LLamafactory framework \cite{zheng2024llamafactory} and LoRA to accelerate the fine-tuning \cite{hu2021loralowrankadaptationlarge}. We train 5 epochs with a GPU of A6000 for 2-3 hours. The rank of LoRA is 8, and the learning rate is 0.0001. We try multiple combinations of hyperparameters before we choose the final hyperparameter set. All the experiments are single runs.

We find out the reasoning without fine-tuning has several issues. The first is inconsistency, for example, the reasoning mentioned the usage of alcohol multiple times, but the final answer fail to include values related to alcohol. Another reasoning does not mention divorce at all, but the final answer choose some values related to divorce. The second is logical errors, for example: \textit{``Oliver's preference for a traditional family setup for raising kids matches When jobs are scarce, employers should give priority to people of this country over immigrants.--Neither agree nor disagree.''} There is no direct logic between two statements. The third is overlooking details. For example, the one-shot reasoning is \textit{``The discussion also involves ethical considerations regarding the use of animals and the environment, which is reflected in the option: Hunting animals as a sport should be banned. --Disagree.''} The reasoning itself sounds reasonable, however, the gold label is \textit{``Using animals for entertainment like in zoos/circuses is ethical.--Disagree.''} The SFT reasoning is \textit{``Finally, Harold’s statement on \textbf{using animals for entertainment}}, which he finds shocking, aligns with the value that such practices are unethical. The difference between the two reasoningg  is that the latter one identified details of \textbf{using animals for entertainment}, while the former one only identified the use of animals and the environment.

\subsection{GRPO Fine-Tuning}
We apply GRPO to fine-tune smaller LLMs on both attitude detection and value selection tasks. For the attitude detection task, which is formulated as multiple-choice classification, the reward signal is binary—1 for correct predictions and 0 otherwise. In contrast, the value selection task is a multi-label setting, where predictions can be partially correct. To account for this, we define the reward as the ratio of true positives to the total number of predicted values. If the number of selected values differs from the ground truth, the reward is set to 0.

Since partial rewards tend to slow convergence compared to binary signals, we train models until performance stabilizes. Empirically, we find that five epochs are sufficient for the attitude detection task, while value selection typically requires around fifteen epochs to converge. We set the rollout number to 16 to ensure stable and fair gradient updates from group-based advantage estimation. We train the model with 4 A100 for 2-3 hours. 
\subsection{Ablation Study}

We conduct ablation studies using the \textbf{Llama~3.2~3B} model on both the \textit{Attitude Detection (AD)} and \textit{Value Selection (VS)} tasks. 
We fine-tune the model with \{50, 100, 250, 500\} stories, both with and without chain-of-thought (CoT) reasoning. 
Each configuration is trained with three random seeds, and we report the mean and standard deviation. 
To examine model size effects, we also evaluate the \textbf{Qwen~2.5~14B} model on the VS task. 

\begin{table*}[h!]
\centering

\scalebox{0.9}{
\begin{tabular}{lcccccc}
\toprule
\textbf{Stories} & \textbf{Setting} & \textbf{Religious} & \textbf{Social} & \textbf{Ethical} & \textbf{Political} & \textbf{Multiple} \\
\midrule
\multirow{2}{*}{50}  
& W/O CoT & 0.5474 (0.0061) & 0.6369 (0.0058) & 0.5025 (0.0126) & 0.6965 (0.0095) & 0.5835 (0.0043) \\
& CoT     & 0.5537 (0.0156) & 0.6761 (0.0101) & 0.5074 (0.0212) & 0.6910 (0.0141) & 0.5729 (0.0072) \\
\midrule
\multirow{2}{*}{100} 
& W/O CoT & 0.6737 (0.0062) & 0.6651 (0.0089) & 0.6655 (0.0046) & 0.6921 (0.0063) & 0.5903 (0.0060) \\
& CoT     & 0.6510 (0.0125) & 0.6636 (0.0059) & 0.6308 (0.0122) & 0.6855 (0.0063) & 0.5714 (0.0105) \\
\midrule
\multirow{2}{*}{250} 
& W/O CoT & 0.6792 (0.0099) & 0.6917 (0.0059) & 0.6951 (0.0182) & 0.6955 (0.0042) & 0.5617 (0.0117) \\
& CoT     & 0.6463 (0.0111) & 0.6479 (0.0102) & 0.6383 (0.0257) & 0.6689 (0.0110) & 0.5615 (0.0052) \\
\midrule
\multirow{2}{*}{500} 
& W/O CoT & 0.6863 (0.0073) & 0.7232 (0.0042) & 0.6716 (0.0087) & 0.6885 (0.0022) & 0.5632 (0.0035) \\
& CoT     & 0.6820 (0.0190) & 0.6918 (0.0079) & 0.6728 (0.0076) & 0.7178 (0.0142) & 0.5623 (0.0065) \\
\bottomrule
\end{tabular}
}
\caption{Ablation study on Attitude Detection using Llama~3.2~3B. We report average accuracy and standard deviation across three seeds.}
\label{tab:ablation_ad}
\end{table*}

\begin{table*}[h!]
\centering

\scalebox{0.9}{
\begin{tabular}{lccccc}
\toprule
\textbf{Stories} & \textbf{Setting} & \textbf{Religious} & \textbf{Social} & \textbf{Ethical} & \textbf{Political} \\
\midrule
\multirow{2}{*}{50}  
& W/O CoT & 0.3204 (0.0070) & 0.3673 (0.0177) & 0.4355 (0.0132) & 0.3571 (0.0339) \\
& CoT     & 0.2849 (0.0035) & 0.2921 (0.0145) & 0.3326 (0.0074) & 0.3116 (0.0116) \\
\midrule
\multirow{2}{*}{100} 
& W/O CoT & 0.3312 (0.0082) & 0.4515 (0.0065) & 0.5080 (0.0666) & 0.4635 (0.0147) \\
& CoT     & 0.3606 (0.0359) & 0.4249 (0.0191) & 0.5003 (0.0264) & 0.3837 (0.0016) \\
\midrule
\multirow{2}{*}{250} 
& W/O CoT & 0.4668 (0.0263) & 0.5429 (0.0119) & 0.6525 (0.0100) & 0.5666 (0.0057) \\
& CoT     & 0.4172 (0.0128) & 0.5300 (0.0051) & 0.6304 (0.0132) & 0.5263 (0.0271) \\
\midrule
\multirow{2}{*}{500} 
& W/O CoT & 0.4881 (0.0113) & 0.5859 (0.0134) & 0.7076 (0.0147) & 0.5957 (0.0110) \\
& CoT     & 0.4036 (0.0059) & 0.5672 (0.0170) & 0.6089 (0.0100) & 0.5181 (0.0029) \\
\bottomrule
\end{tabular}
}
\caption{Ablation study on Value Selection using Llama~3.2~3B. We report average accuracy and standard deviation across three seeds.}
\label{tab:ablation_vs_llama}
\end{table*}

\begin{table*}[h!]
\centering

\scalebox{0.9}{
\begin{tabular}{lccccc}
\toprule
\textbf{Stories} & \textbf{Setting} & \textbf{Religious} & \textbf{Social} & \textbf{Ethical} & \textbf{Political} \\
\midrule
\multirow{2}{*}{50}  
& W/O CoT & 0.4272 (0.0156) & 0.6093 (0.0087) & 0.6916 (0.0079) & 0.5625 (0.0154) \\
& CoT     & 0.4642 (0.0091) & 0.5971 (0.0179) & 0.6758 (0.0093) & 0.5996 (0.0021) \\
\midrule
\multirow{2}{*}{100} 
& W/O CoT & 0.4210 (0.0167) & 0.5909 (0.0067) & 0.6902 (0.0179) & 0.5708 (0.0062) \\
& CoT     & 0.4951 (0.0109) & 0.5876 (0.0101) & 0.6720 (0.0137) & 0.5634 (0.0251) \\
\midrule
\multirow{2}{*}{250} 
& W/O CoT & 0.4099 (0.0157) & 0.6002 (0.0060) & 0.6794 (0.0117) & 0.5821 (0.0149) \\
& CoT     & 0.5461 (0.0076) & 0.5868 (0.0148) & 0.6697 (0.0148) & 0.5962 (0.0082) \\
\midrule
\multirow{2}{*}{500} 
& W/O CoT & 0.4817 (0.0069) & 0.6129 (0.0067) & 0.7672 (0.0043) & 0.6663 (0.0032) \\
& CoT     & 0.5465 (0.0122) & 0.6991 (0.0148) & 0.7524 (0.0200) & 0.6935 (0.0039) \\
\bottomrule
\end{tabular}
}
\caption{Value Selection results on Qwen~2.5~14B.}
\label{tab:ablation_vs_qwen}
\end{table*}

We summarize key findings as follows:

\begin{itemize}
    \item \textbf{Impact of dataset size.} In attitude detection, increasing dataset size does not strongly correlate with performance, likely because the task involves only two to three attitude options and the pattern is quickly learnable. In contrast, value selection shows consistent improvement as dataset size increases, especially from 50 to 250 samples. 
    \item \textbf{Effect of CoT reasoning.} Surprisingly, non-reasoning (W/O CoT) fine-tuning often outperforms CoT in the Llama~3.2~3B model, particularly in the VS task. Manual inspection reveals that smaller models struggle to maintain coherent reasoning chains, often producing disorganized outputs.
    \item \textbf{Effect of model size.} For the Qwen~14B model, CoT begins to show advantage as dataset size increases, indicating that larger models can better leverage reasoning supervision. While the improvements are not universally significant, results suggest that CoT becomes beneficial only beyond a certain data scale and model capacity.
    \item \textbf{Alignment with prior findings.} Our observations are consistent with prior work \citep{kojima2023cotnotcot}, which shows that CoT mainly benefits symbolic or mathematical reasoning tasks, and may not generalize as effectively to social or value-based reasoning.
\end{itemize}

\subsection{Case study on value expression}
To better assess the model’s ability to express values implicitly, we conduct a small case study using Qwen 2.5 7B. The model is prompted to respond to an ongoing conversation given a target value, with the goal of reflecting the value implicitly while maintaining natural conversational flow. We compare the vanilla Qwen 2.5 7B model with its GRPO-tuned counterpart.

To evaluate the quality of generated speech, we employ an LLM-as-a-judge framework that assesses three dimensions: value reflection, speech naturalness, and value implicitness. We experiment with two judging strategies: (1) pairwise comparison, where the LLM directly compares two speeches, and (2) independent rating, where it scores each speech separately before comparison.
Across both evaluation methods, the GRPO-tuned model performs slightly better, with approximately 5\% of speeches showing improvement, though the gain is not statistically significant. This case study suggests that while GRPO tuning enhances implicit value expression modestly, further improvements could be achieved by jointly strengthening detection and expression capabilities.

\section{Human Annotations}

To compare the human performance with LLMs, we ask several human participants from CS undergraduates and master's students. We cover students from different background including 2 Chinese students and 2 Indian students. We provide the exact prompt we use for LLM to those students.


\clearpage
\section{Prompts}
In this section, we show all the prompts we use for story generation, experiments, and validation. 

\begin{tcolorbox}[colback=blue!10!white, colframe=blue!50!black, title=Story generation prompt,fonttitle=\bfseries]
\#\#Identity

You are an expert in culture understanding.

\#\# Task

You will be provided with 5 cultural values and a location where the conversation happens. Each of them follows the format [culture]--[value]. The first [culture] describes a statement or a situation, and [value] is how you agree with the culture or is the culture common or not.

Your task is to generate a scene including conversations and actions among multiple people and the scene needs to reflect the culture values provided.
   
   Here are some requirements of the scene:
   
   1. It cannot be too short. It should have multiple rounds of interaction among people.
   
   2. It should not be too obvious. It cannot directly spit out or rephrase the values.
   
   3. It cannot be easy to human to understand the culture values behind.
   
   4. You do not need to follow the order of the values. You could mention the values multiple times through the conversation. Make sure the conversation flows well.  
   
   5. All the characters should follow the given values. There should not be contradictions between the character's value and the given value.

\#\#Input

   Here are the cultural values you should follow when generating:
   {values}

   Here is the pre-defined location of the scene:
   {location}

\#\#Output
   Now using the cultural values to generate a story.
\end{tcolorbox}%

\begin{tcolorbox}[colback=blue!10!white, colframe=blue!50!black, title=Incorporation check prompt,fonttitle=\bfseries]
\#\#Identity

You are an expert in culture understanding.

\#\#Task

You will be provided a story and values reflected in the story.
Your task is to check if the story reflects values? The story does not mention the values directly. 
You will need some reasoning to analyze the story.

\#\#Input

Here is the original story:
{story}

Here are the values to reflect in the story:
{values}

\#\#Output

For each value, output if the value is reflected and provide reasoning.
In the end, output the values not reflected without the reasoning. 
Only output the exact and comprehensive value including "--" within it, do not rephrase! If all the values are reflected, just leave it blank. 
Follow the format:
[Value]:[Reasoning, Yes/No]
.....
Values not reflected:
[Value]

\end{tcolorbox}

\begin{tcolorbox}[colback=blue!10!white, colframe=blue!50!black, title=Missing values incorporation prompt,fonttitle=\bfseries]
\#\#Identity

You are an expert in culture understanding.

\#\#Task

You will be provided a story and values which need to be reflected in the story.

Your task is to refine the story to reflect the value provided. 
You cannot remove anything or replace existing speeches from the story, you can only add conversations to reflect the value.

The refinement should flow with original story well. You cannot add new conversation randomly.

\#\#Input
Here is the original story:
{story}

Here are the values to reflect in the story:
{values}

\#\#Output
Now refine the story.

\end{tcolorbox}

\begin{tcolorbox}[colback=blue!10!white, colframe=blue!50!black, title=Consistency check prompt,fonttitle=\bfseries]
\#\#Identity

You are an expert in culture understanding.

\#\#Task
You will be provided a story and values reflected in the story.
Your task: for each value, check if all the characters agree with the value. If there is characters who does not agree with the value, you should output the character's name and his speech, and why the speech does not align with the value.

\#\#Input
Here is the original story:
{story}

Here are the values to reflect in the story:
{values}

Now check the story and output if there is any contradiction. You can output reasoning to help you analyze. However, in the end, only output where is the contradiction one by one. If there isn't a contradiction, just reply NO. Otherwise, reply where is the contradiction.

\#\#Output
Follow the format strictly, do not change the format, output exact values from the values provided, and do not rephrase:

[Reasoning]:

[Value--attitude]:[If all the speeches are aligned with the value]

[Contradictions]:

[Value--attitude]: [*character name*:speech]
......

\end{tcolorbox}

\begin{tcolorbox}[colback=blue!10!white, colframe=blue!50!black, title=Consistency resolve prompt,fonttitle=\bfseries]
\#\#Identity

You are an expert in culture understanding.

\#\#Task

You will be provided a story and values which need to be reflected in the story.
However, the story includes some contradiction where characters do not agree on certain values. You will be provided where is the contradiction.

The contradiction includes 3 parts: 

1. Correct value to follow

2. Character name

3. Character speech

Your task is to replace the speech mentioned in the contradictions with a new speech to make sure the speech is aligned with the values
The refinement should flow with the original story well. You cannot add new conversations randomly.

\#\#Input
Here is the original story:
{story}

Here are the contradictions:
{contradiction}

Ignore the original character's speech. Directly write a new speech that reflects the value.

\#\#Output
Here is the rewritten speech:

\end{tcolorbox}

\begin{tcolorbox}[colback=blue!10!white, colframe=blue!50!black, title=Implicitness check prompt,fonttitle=\bfseries]
\#\#Identity

You are an expert in culture understanding.

\#\#Task
You will be provided a story and values reflected in the story.
Your task is to check if there is obvious speech that directly mentions or rephrases the values. If the story mentions phrases or sentences from values, that would be also counted as directly mentioned
If it just reflects the value but does not rephrase the values, it is not considered as obvious speech.
Here are some examples of obvious speech:

[Example 1]

Value: In your view, how often do the following things occur in this country’s elections: Journalists provide fair coverage of elections—Very often
Speech: "I think journalists do their best to cover things fairly"
Explanation: The speech basically rephrases the value without providing supporting evidence

[Example 2]:
.......

You should output places where the values are directly mentioned. If the values are not directly spit out, do not output them!!! 

\#\#Input
Here is the original story:
{story}

Here are the values to reflect in the story:
{values}

\#\#Output
Now check if any parts of the stories directly mentioned the values. You can output reasoning or evidence and help you analyze.
However, in the final answer, output speeches where the values are directly mentioned and the corresponding values. If there are no directly mentioned values, just output NO in the final answer.

Follow the format:

[Reasoning]: reasoning

[Final answer]: [value]:[speech]

[value]:[speech]
.....

\end{tcolorbox}

\begin{tcolorbox}[colback=blue!10!white, colframe=blue!50!black, title=Implicitness rewrite prompt,fonttitle=\bfseries]
\#\#Identity

You are an expert in culture understanding.

\#\#Task

You will be provided a story and values reflected in the story.
You will be also provided with some comments where the values are directly mentioned in the story.
Each comment includes a value and a speech where the speech directly reflects the value.
Your task is to refine the speech by rewriting the places where they directly mention the values.
Instead of speaking out values directly, characters should reflect values implicitly by using anecdotes.
Imagine the place is not a safe place to speak out, characters need to respect each other and not offend others.
You need to make sure after rewriting, people are able to detect values by the speech. It cannot be too abstract.

Here are some examples:

[Example 1]:
Value:  How frequently do the following things occur in your neighborhood: Robberies?--Not frequently
Original speech: “Well, at least we’re not dealing with robberies or anything in our area.” 
Refined speech: "You know in other counties, cars got breaking in is common. At least we don't have to worry about that."

[Example 2]:
.....

\#\#Input
Here is the original story:
{story}

Here are the values to reflect in the story:
{values}

Here are the comments that you need to refer to when refining the story:
{comments}

\#\#Output

Please rewrite the speech one by one in the story. Do not miss any comments.

For each rewritten speech, you need to follow this format:
Value:[value]
Original speech: [Speech in the comment]
Refined speech: [You rewritten speech]

Here is your output:
\end{tcolorbox}

\begin{tcolorbox}[colback=blue!10!white, colframe=blue!50!black, title=Attitude detection prompt,fonttitle=\bfseries]
\#\#Identity

You are an expert in culture understanding.

\#\#Task

You will be given a scenario including background, conversation and actions in a location. You task is to detect the culture values reflected from the conversation.
You will be provided a statement and you need to detect what attitude they hold to the value.
For example, the statement is "The people who don't work turn lazy" and the options are "disagree, agree, neither disagree nor agree". And you need to read the story and choose one attitude from the options.
The options could also describe something happening around them. For example, if the statement is "How often does this happen in your neighborhood:Racist behaviors", the options are "Very often, not often, not at all often", you need to choose one based on the story.

\#\#Input
Here is the scenario:
{story}

Here is the statement:
{statement}

Here are the options:
{options}

\#\#Output
Now choose one option based on the story. Let's think step by step! First, extract the speech that is related to the statement.Then analyse the attitude of those people from the speech. Finally, output the answer.

Follow the format:

[Related speech]: speeches

[Analysis]: analysis of attitude

[Answer]:answer
\end{tcolorbox}

\begin{tcolorbox}[colback=blue!10!white, colframe=blue!50!black, title=Value selection prompt,fonttitle=\bfseries]
\#\#Identity

You are an expert in culture understanding.

\#\#Task
You will be given a scenario including background, conversation and actions in a location. You task is to detect the culture values reflected from the conversation.

You will be provided 15 options and you need to select {number} correct answers from the options. (Only {number}  options are correct).

You need to choose the whole option, for example:

Being a housewife is just as fulfilling as working for pay--agree

Make sure you include '--' in the answer.

\#\#Input

Here is the scenario:
{story}

Here are the options:
{options}

\#\# Instruction
To think step by step:
1. You need to first detect what topics the story mentioned according to the options. You need to list the speech related to the topic. 
2. Then for each topic, detect what values might be related to the topic. Sometimes, you might find multiple options that might be correct, for uncertain options, you should compare them. 
3. Based on the previous detection, analyze what values are aligned with the story, you should always prioritize those values that are strongly related. 
4. In the end, output the final answers only. Only choose the required number of values.

\#\#Output
Follow the format:
[Topic]: 
topic1: speeches related to the topic
.....

[Value detection]: 
topic1: what values are related to the topic

[Reasoning]:
Reasoning

[Final answer]: 

[text] -- [text] 

.....

"""
\end{tcolorbox}

\begin{tcolorbox}[colback=blue!10!white, colframe=blue!50!black, title=Value extraction prompt,fonttitle=\bfseries]
\#\#Identity

You are an expert in culture understanding.

\#\#Task

You will be provided with a story. Your goal is to identify the most prevalent {topic} values in the story.
    For example, one religious value could be: I strongly believe in God and the afterlife. 
    The value should be a complete sentence, it should not be a phrase like work-life balance.
    The value should hold a attitude or it reflects social phenomena, it should not be a overview of topic. 
    For example, one social value could be: I think work is a duty towards the society.
    And one political value could be: I think voters are bribed in our election system.
    Please identify the values in this story: {story} by paying attention to how the characters in the story discuss {topic}.
    You need to generate 10 values in total, make sure values are specific and detailed. Only focus on one topic in one value. Do not mention several themes in one value.

\#\#Input
    Here is the story: {story} 
    
    Here is a story summary with some key ideas: {summary}. 
    
\#\#Output

    You could output reasoning before you output final answer. But in the end, your output should follow the format:
    [Final answer]:
    value1
    value2
    ....
\end{tcolorbox}

\begin{tcolorbox}[colback=blue!10!white, colframe=blue!50!black, title=LLM evaluation prompt,fonttitle=\bfseries]
\#\#Identity

You are an expert in culture understanding.

\#\#Task

You will be given two sets of texts: a set of predicted values and a ground truth set of values.
   Your task is to determine how many of the ground truth values are fully represented in the predicted values.
   A ground truth value is considered correct if all of its components are meaningfully discussed in the predicted value, even if there is no exact 1-to-1 match. It could be many-to-1 match i.e. many values to 1 ground truth value.
   If the ground truth value is fully presented, score 1, if it is partially presented, score 0.5, if it is not mentioned at all, score 0.
   Additionally, provide a brief justification for your score, explaining which values were correctly or incorrectly represented, in the justification, you should explicitly mention which predicted values are related to the ground truth value.
   
\#\#Input

   Here are the predictions: {pred} and ground truth: {gt}
   In the reasoning, if the ground truth value is fully represented, you need to point out which predicted value is related to it.

\#\#Output

   Return the results in this format:
   [Reasoning]:
   
   [Ground truth 1]: [reasoning]
   
   .....
   
   [Final answer]:
   
   [Ground truth 1]:1
   
   [Ground truth 2]:0.5
   
   ......

\end{tcolorbox}

\end{document}